\documentclass[sigconf, 10pt]{acmart}
\usepackage{xcolor}
\usepackage{xurl}
\usepackage{booktabs}
\usepackage[ruled]{algorithm2e}
\usepackage{algpseudocode}
\usepackage{enumerate}
\usepackage[english]{babel}
\usepackage{blindtext}
\usepackage{amsmath}

\usepackage{amssymb}
\usepackage{xspace}
\usepackage{multirow}
\usepackage{threeparttable}
\usepackage{float}
\usepackage{flafter}
\usepackage{comment}
\usepackage[skip=12pt]{caption}
\usepackage{graphicx}
\usepackage{subcaption}
\usepackage{adjustbox}
\usepackage{caption}
\usepackage[per-mode=symbol]{siunitx}

\def\fig{Fig.\xspace}

\def\tab{Tab.\xspace}

\newcommand{\head}[1]{{\noindent \textbf{#1:}}}

\graphicspath{{./figures/}}
\graphicspath{{./pdf/}}
\DeclareGraphicsExtensions{.pdf,.jpeg,.png}

\renewcommand\footnotetextcopyrightpermission[1]{}
\setcopyright{none}
\acmDOI{}

\acmISBN{}

\acmPrice{}
\def\sysname{HARGPT\xspace}
\begin{document}
\title{\sysname: Are LLMs Zero-Shot Human Activity Recognizers?
}

\author[Sijie Ji, Xinzhe Zheng, Chenshu Wu]{Sijie Ji$^{*}$, Xinzhe Zheng$^{*}$ , Chenshu Wu \thanks{Sijie Ji and Xinzhe Zheng are the co-first authors.}}
\affiliation{%
  \institution{The University of Hong Kong}
  \country{Email:\{sijieji, xzzheng, chenshu\}@cs.hku.hk}
}
\def \authors{Sijie Ji, Xinzhe Zheng, Chenshu Wu}

\renewcommand{\shortauthors}{\sysname}

\begin{abstract}
There is an ongoing debate regarding the potential of Large Language Models (LLMs) as foundational models seamlessly integrated with Cyber-Physical Systems (CPS) for interpreting the physical world. 
In this paper, we carry out a case study to answer the following question: Are LLMs capable of zero-shot human activity recognition (HAR)? 
Our study, HARGPT, presents an affirmative answer by demonstrating that LLMs can comprehend raw IMU data and perform HAR tasks in a zero-shot manner, with only appropriate prompts. 
HARGPT inputs raw IMU data into LLMs and utilizes the \textit{role-play} and \textit{``think step-by-step''} strategies for prompting. 
We benchmark HARGPT on GPT4 using two public datasets of different inter-class similarities and compare various baselines both based on traditional machine learning and state-of-the-art deep classification models. 
Remarkably, LLMs successfully recognize human activities from raw IMU data and consistently outperform all the baselines on both datasets. 
Our findings indicate that by effective prompting, LLMs can interpret raw IMU data based on their knowledge base, possessing a promising potential to analyze raw sensor data of the physical world effectively.
\end{abstract}

\pagestyle{plain} 
\maketitle

\section{Introduction}
\label{sec:intro}

Recent advancements in Foundation Models (FMs), particularly Large Language Models (LLMs) and Large Multimodal Models (LMMs), have garnered significant attention due to their remarkable capabilities in efficiently handling a diverse array of downstream tasks. 
Foundation models are noteworthy for two primary reasons: their ability to mimic human reasoning at a high level and their exceptional generalization abilities.
Most importantly, some evidence shows that LLMs have an emergent ability to understand the physical world in a manner akin to that of a child.
This is exemplified by Sora~\cite{videoworldsimulators2024}, which exhibits intuitive comprehension of various aspects of the world, ranging from fundamental physics principles to complex societal and artistic concepts.

\begin{figure}[t]
  \begin{center}
  \includegraphics[width=0.46\textwidth]{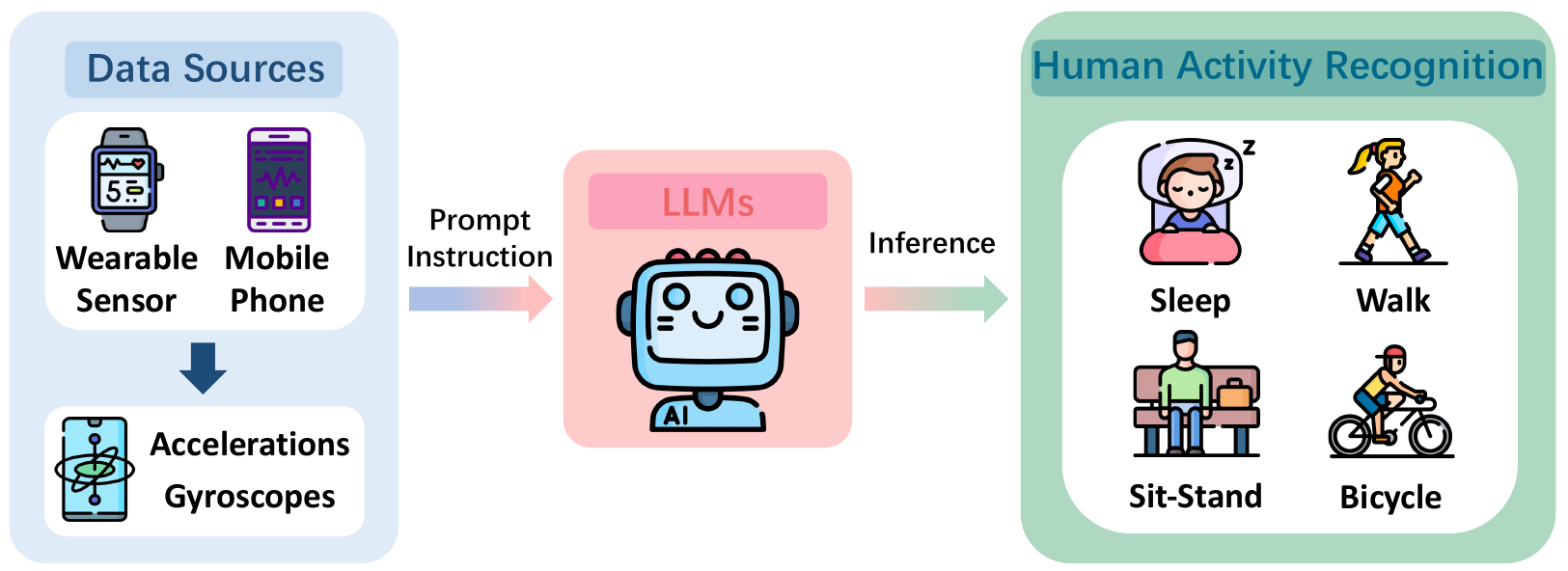}
  \caption{Workflow of HARGPT.}\label{fig:hargpt-workflow}
  \end{center}
  \vspace{-0.5cm}
\end{figure}

Despite these impressive feats, some researchers posit that current LLMs possess only surface-level knowledge and are still far from achieving a deep understanding of the physical world~\cite{lecun2022path}.
This limitation is attributed to their training on vast amounts of internet-scale text corpora and images, leading to subpar performance when analyzing digital and time series data.
Additionally, these models often struggle when attempting to interact with the physical world in a meaningful way, such as generating precise control sequences~\cite{li2024personal}.

This ongoing debate has sparked considerable interest in exploring the potential of LLMs to serve as foundational models seamlessly integrated with Cyber-Physical Systems (CPS) for interpreting the physical world.
In particular, researchers propose the concept of Penetrative AI and explore the integration of LLM with the physical world from two levels: textual signals and raw digital signals~\cite{xu2024penetrative}.
However, the cases they show, such as inferring user location through WiFi signals, mainly rely on explicitly direct textural information and they require specific prompt engineering to provide expert knowledge about the usage of sensor data.
The true potential of LLMs as a physical world model remains largely unexplored.

\begin{figure*}[t]
   \subfloat[Capture24]{%
    \label{fig:capture24v}
      \includegraphics[width=0.5\textwidth]{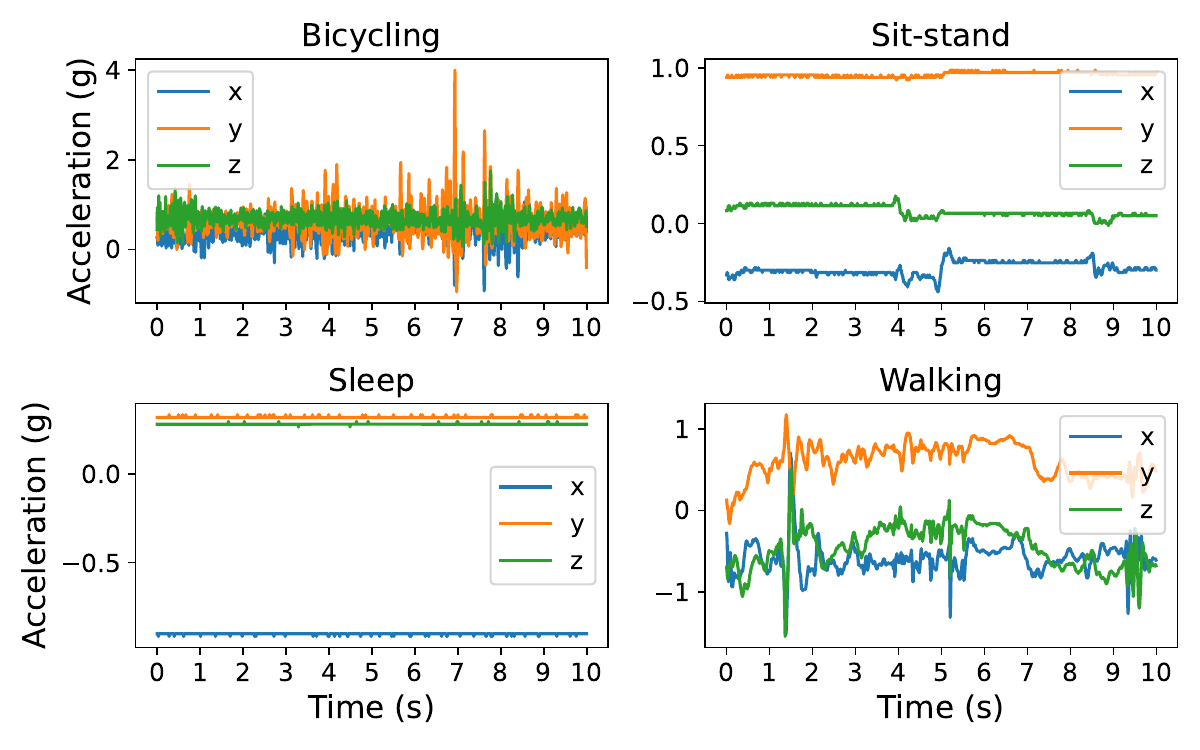}
	  }%
   \subfloat[HHAR]{%
    \label{fig:hharv}
      \includegraphics[width=0.25\textwidth]{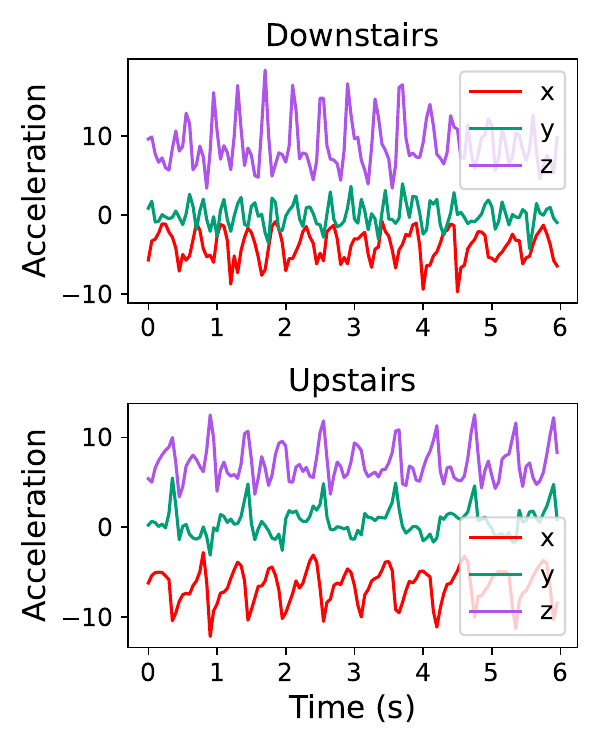}
	  }%
   \caption{IMU data visualization of two datasets. (a): Capture24 dataset contains four HAR categories with distinct patterns; (b): HHAR dataset contains two similar HAR categories.}
\end{figure*}

Motivated by the advancements of LLMs and the potential of Penetrative AI, this paper starts from the classic CPS application Human Activity Recognition (HAR), aiming to explore and understand the current capabilities of LLMs by directly inputting raw sensor data with a simple prompt. 
As illustrated in \fig\ref{fig:hargpt-workflow}, the raw IoT sensor data is fed directly into various well-known LLMs such as ChatGPT~\cite{kocon2023chatgpt}, Google Gemini~\cite{team2023gemini}, and LLaMA2-70b~\cite{touvron2023llama}, yielding recognition outcomes.
The performance evaluation on two distinct benchmark datasets with two levels of difficulty, one with distinct patterns, such as sleeping and walking, and one with inter-class similarities, such as climbing upstairs and downstairs. 
The results of the experiments demonstrate that LLMs are capable of performing zero-shot HAR using raw sensor data, achieving an average accuracy of 80\% on unseen data (same activity performed by unseen users), which outperforms existing methods, either for classic machine learning or deep learning methods. 
More importantly, unlike conventional approaches that are prone to performance degradation when confronted with unseen data and often necessitate retraining or fine-tuning for specific datasets, LLMs exhibit a high degree of robustness. In summary, we present HARGPT and make the following remarks:

\begin{itemize}
    \item 
    HARGPT first discovers that LLMs can function as zero-shot human activity recognizers without the need for fine-tuning or domain-specific expertise-guided prompt engineering.
    \item HARGPT showcases the proficiency of LLMs in processing IoT sensor data and carrying out tasks in the physical world.
\end{itemize}
Furthermore, we delve into exploration and discourse on the insights gained and notable discoveries made while utilizing LLMs for handling IoT sensor data, paving the way for future research on adopting LLMs for CPS.

\section{HARGPT: zero-shot HAR with LLMs}
\label{sec:design}
\subsection{Experiments}
To evaluate the capability of LLMs for HAR based on IMU raw data, we conducted two levels of experiments varying in difficulty.
The initial set of experiments aimed to determine if LLMs could differentiate between movement types with distinct patterns, such as sleeping and walking, which exhibit obvious dissimilarities.
In the subsequent set of experiments, we sought to ascertain whether these models could further discriminate between movements that possess highly similar patterns, such as upstairs and downstairs, which are challenging to differentiate even for human observers.

\subsubsection{Dataset Setup}
To facilitate these experiments, we utilized two distinct datasets: Capture24~\cite{chan2021capture} and HHAR~\cite{stisen2015smart}, for the experiments respectively.
Below, further details regarding each dataset are provided.

\head{Capture24}
Capture24 is prepared for the first experiment.
The dataset contains human motion activities in daily living with accelerations only.
The data is collected from wristwatches with a sampling rate of 100Hz.
For the experimental task, we only use the IMU data with labels of sleep, walking, bicycle, and sit-stand.
The visualization outcome is given in \fig\ref{fig:capture24v}.
These categories of data have significantly distinct characteristics and are relatively easy to distinguish.

To compare LLMs with the baselines that need training data, we partition the dataset by the user into training, validation, test seen, and test unseen datasets at a ratio of 4:1:1:2.
Additionally, given the requirement to input raw IMU data as tokens into LLMs, we down-sample the IMU data to 10Hz for LLMs.

\head{HHAR}
HHAR is utilized for the second experiment, so we evaluate downstairs and upstairs data exclusively.
It contains readings from accelerometers and gyroscopes from 9 users.
The data is collected from mobile phones carried by the users around their waists with sampling rates of 100 and 200Hz.
We visualize the IMU data in \fig\ref{fig:hharv}, which further confirms our assumptions.

HHAR is partitioned by the user into training, validation, test seen, and test unseen datasets at a ratio of 4:1:1:2 as well.
The IMU data is down-sampled to 10Hz as tokens for LLMs.

\subsubsection{Baseline}
Classical machine learning models and state-of-the-art deep learning models for HAR are taken as baseline models.
These models need to be trained to acquire classification ability.

\head{Random Forest~\cite{biau2016random}}
Random forests (RF) is an ensemble learning method for classification.
It operates by constructing a multitude of decision trees at training time.
With its specific learning strategy, RF could handle minor classes well in the classification tasks.
For implementation, we use the model provided in scikit-learn~\cite{scikit-learn} with its default settings.

\head{SVM~\cite{hearst1998support}}
support vector machine (SVM) is a supervised max-margin model with associated learning algorithms that analyzes data for classification and regression.
We use the SVM model in scikit-learn with Gaussian kernels.

\head{DCNN~\cite{yang2015deep}}
DCNN is composed of convolutional neural networks that can automatically learn features from multichannel time series signals acquired from body-worn sensors for HAR.
It can learn complex HAR features, and achieve promising classification results compared with some traditional methods, like deep belief network.

\head{LIMU-LSTM~\cite{xu2022limu}}
LIMU-LSTM is a classification model constructed with long short-term memory (LSTM).
Different from other baseline models, LIMU-LSTM analyzes the input features in chronological order.

\subsubsection{Prompt Structure}
LLMs are generally known as excellent few-shot learners, where one can use a text or template known as a prompt to strongly guide the generation to output answers for desired tasks~\cite{wei2022chain}. For example, in a recent study~\cite{nori2023can}, researchers utilize a guiding instruction to release the professional medical domain knowledge of LLMs, achieving state-of-the-art performance across all benchmark datasets. The problem with such prompt engineering is that they need to manually construct the question-and-answer template, which makes the LLMs restricted to specific downstream applications~\cite{xu2024penetrative}. Further, the study also shows that such a hand-crafted prompting template will go against the Chain-of-Thought property of LLMs and result in bad performance~\cite{kojima2022large}. 
To circumvent the above problems, we structure our prompt in the simplest way, using role-play instructions and let's think step-by-step guidance to directly trigger and demonstrate the original capabilities of LLMs without giving an answer template. 

\begin{figure}[t]
  \begin{center}
  \includegraphics[width=0.45\textwidth]{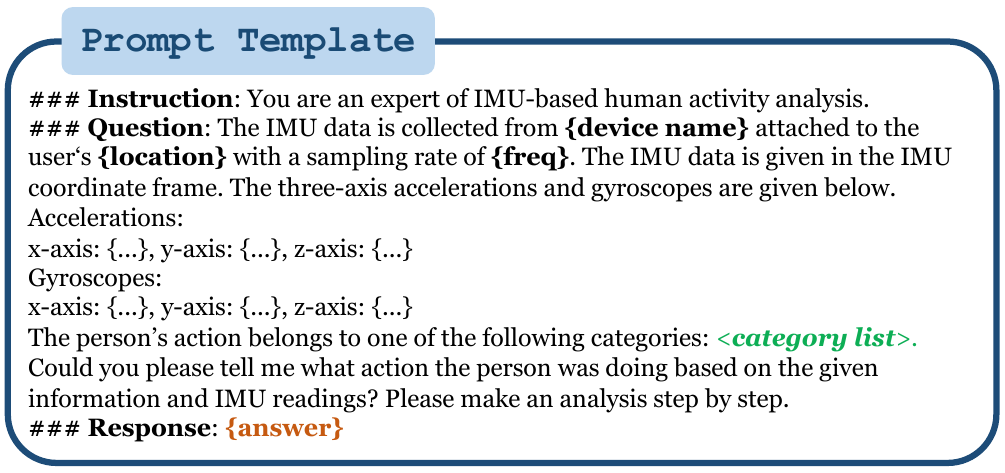}
  \caption{Chain-of-thought prompt design for HARGPT.}\label{fig:prompt}
  \end{center}
\end{figure}

Our prompt design, depicted in \fig\ref{fig:prompt}, comprises only an instruction and a question.
The instruction aims to leverage the expert knowledge of LLMs regarding IMU.
The question provides specific details about data collection, down-sampled raw data sequence, and potential categories of human actions.
By concluding the question with the phrase "Please make an analysis step by step," we aim to elicit a detailed chain-of-thought (CoT) process from LLMs, as this approach has been proven to enhance the accuracy of their answers in existing literature~\cite{kim2024health,wei2022chain}. As shown in \fig\ref{fig:gpt4inference}, when we do not restrict the answer template of LLM, it will generate richer text information utilizing the inherent reasoning capability to retrieve the corresponding embedding knowledge.

It should be noted that Capture24 only contains acceleration data, so we exclude gyroscope information during testing.
Additionally, we conduct a comparative test to assess the impact of CoT on improving the accuracy of question answering.
In the experiment, the last sentence is replaced with "\textit{Please give your answer directly without analysis.}"

\subsection{Evaluation}
We select GPT4~\cite{achiam2023gpt}, the most advanced and powerful LLM currently accessible, to conduct a comprehensive analysis with four other baseline models across two datasets.
Additionally, we compare the prompt mode of direct output (DO) without analysis to verify the effectiveness of applying CoT prompt for LLMs to improve prediction accuracy.

\head{Inter-class Difference}
We first test GPT4 and other baseline models on the Capture24 dataset, which contains four generally distinct HAR categories, including sleep, walking, sit-stand, and bicycling.
The overall testing result is shown in \tab\ref{tab:res-capture24}.
Undoubtedly, leveraging its robust comprehension capabilities and CoT prompt, GPT4 has exhibited exceptional performance across all baseline measures on the unseen set, boasting an average f1-score of 0.795.
Specifically, when compared with GPT4-DO, GPT4-CoT demonstrates a noteworthy improvement of 0.33.
In contrast, even the best baseline DCNN achieves only a modest level of approximately 0.6.
LIMU-LSTM marginally underperforms compared to DCNN.
Conversely, the two remaining machine learning methods exhibit poor performance.
For instance, the classification results of each category are presented in \tab\ref{tab:per-capture24}, and it becomes evident that GPT4-CoT could achieve a well-balanced performance.
Notably, while DCNN and LIMU-LSTM exhibit high accuracy in predicting bicycling and sleep activities, their accuracy significantly diminishes when predicting sit-stand and walking activities.
This disparity in accuracy might be attributed to the increased freedom of motion associated with the latter two activities.

\begin{table}[t]
\resizebox{0.9\columnwidth}{!}{%
\begin{tabular}{@{}c|c|ccc@{}}
\toprule
\multirow{2}{*}{Method}        & \multirow{2}{*}{Test Subject} & \multicolumn{3}{c}{Evaluation Metric   (macro avg.)}                                       \\ \cmidrule(l){3-5} 
                      &        & \multicolumn{1}{c|}{Precision} & \multicolumn{1}{c|}{Recall} & F1-Score \\ \midrule
\multirow{2}{*}{RF} & Seen                          & \multicolumn{1}{c|}{0.560}          & \multicolumn{1}{c|}{0.635}          & 0.580          \\ \cmidrule(l){2-5} 
                      & Unseen & \multicolumn{1}{c|}{0.525}     & \multicolumn{1}{c|}{0.598}  & 0.555    \\ \midrule
\multirow{2}{*}{SVM}  & Seen   & \multicolumn{1}{c|}{0.463}     & \multicolumn{1}{c|}{0.505}  & 0.478    \\ \cmidrule(l){2-5} 
                      & Unseen & \multicolumn{1}{c|}{0.535}     & \multicolumn{1}{c|}{0.598}  & 0.545    \\ \midrule
\multirow{2}{*}{DCNN} & Seen   & \multicolumn{1}{c|}{0.615}     & \multicolumn{1}{c|}{0.628}  & 0.615    \\ \cmidrule(l){2-5} 
                      & Unseen & \multicolumn{1}{c|}{0.595}     & \multicolumn{1}{c|}{0.600}  & 0.588    \\ \midrule
\multirow{2}{*}{LIMU-LSTM} & Seen   & \multicolumn{1}{c|}{0.615}     & \multicolumn{1}{c|}{0.628}  & 0.618    \\ \cmidrule(l){2-5} 
                      & Unseen & \multicolumn{1}{c|}{0.595}     & \multicolumn{1}{c|}{0.588}  & 0.585    \\ \midrule
$\rm{GPT4 - DO^*}$             & Unseen & \multicolumn{1}{c|}{0.498}     & \multicolumn{1}{c|}{0.468}  & 0.465    \\ \midrule
$\rm{GPT4 - CoT^*}$                     & Unseen                        & \multicolumn{1}{c|}{\textbf{0.818}} & \multicolumn{1}{c|}{\textbf{0.793}} & \textbf{0.795} \\ \bottomrule
\end{tabular}%
}
\caption{Overall test results on Capture24 dataset. {(\rm{$\rm{DO^*}$: direct output; $\rm{CoT^*}$: chain-of-thought.)}}}
\label{tab:res-capture24}
\end{table}

\begin{table}[t]
\resizebox{0.9\columnwidth}{!}{%
\begin{tabular}{@{}c|c|c|c|c|c|c|c@{}}
\toprule
Label                      & Metrics   & RF   & SVM  & DCNN          & LIMU-LSTM     & GPT4 - DO & GPT4 - CoT      \\ \midrule
\multirow{3}{*}{Bicycling} & Precision & 0.80 & 0.89 & 0.95          & \textbf{0.97} & 0.53      & 0.64          \\ \cmidrule(l){2-8} 
                           & Recall    & 0.81 & 0.79 & 0.77          & 0.74          & 0.36      & \textbf{0.84} \\ \cmidrule(l){2-8} 
                           & F1-score  & 0.81 & 0.84 & \textbf{0.85} & 0.84          & 0.43      & 0.72          \\ \midrule
\multirow{3}{*}{Sit-stand} & Precision & 0.24 & 0.29 & 0.32          & 0.26          & 0.35      & \textbf{0.89} \\ \cmidrule(l){2-8} 
                           & Recall    & 0.41 & 0.46 & 0.43          & 0.33          & 0.32      & \textbf{0.73} \\ \cmidrule(l){2-8} 
                           & F1-score  & 0.30 & 0.36 & 0.37          & 0.29          & 0.33      & \textbf{0.80} \\ \midrule
\multirow{3}{*}{Sleep}     & Precision & 0.61 & 0.47 & 0.75          & 0.80          & 0.71      & \textbf{1.00} \\ \cmidrule(l){2-8} 
                           & Recall    & 0.75 & 0.83 & \textbf{0.97} & 0.96          & 0.50      & 0.83          \\ \cmidrule(l){2-8} 
                           & F1-score  & 0.67 & 0.60 & 0.85          & 0.88          & 0.59      & \textbf{0.91} \\ \midrule
\multirow{3}{*}{Walking}   & Precision & 0.45 & 0.49 & 0.36          & 0.35          & 0.40      & \textbf{0.74} \\ \cmidrule(l){2-8} 
                           & Recall    & 0.42 & 0.31 & 0.23          & 0.32          & 0.69      & \textbf{0.77} \\ \cmidrule(l){2-8} 
                           & F1-score  & 0.44 & 0.38 & 0.28          & 0.33          & 0.51      & \textbf{0.75} \\ \bottomrule
\end{tabular}%
}
\caption{Performance of each action category on Capture24 dataset.}
\label{tab:per-capture24}
\end{table}

\head{Inter-class Similarity}
To avoid the issue of the obvious pattern difference in the first experiment, we conducted a comparison challenge test, in which the model is asked to discern between two similar actions, specifically ascending and descending stairs.
\fig\ref{fig:hharv} visually depicts the results obtained from this investigation.
The experimental findings are presented in \tab\ref{tab:res-hhar} and \tab\ref{tab:per-hhar}.
Similarly, GPT4-CoT demonstrates outstanding outcomes, exhibiting an average accuracy close to 80\%, along with a commendable recall rate and f1-score.
On the other hand, the baseline models, while capable of achieving high accuracy on test seen set (where three of the baseline models surpass 90\% accuracy), struggle to perform adequately on test unseen samples.
Even the top-performing LIMU-LSTM exhibits a nearly 10\% decrease when compared to GPT4-CoT on the unseen dataset.

\begin{table}[t]
\resizebox{0.9\columnwidth}{!}{%
\begin{tabular}{@{}c|c|ccc@{}}
\toprule
\multirow{2}{*}{Method}    & \multirow{2}{*}{Test Subject} & \multicolumn{3}{c}{Evaluation Metrics   (macro avg.)}                                      \\ \cmidrule(l){3-5} 
                      &        & \multicolumn{1}{c|}{Precision} & \multicolumn{1}{c|}{Recall} & F1-Score \\ \midrule
\multirow{2}{*}{RF}   & Seen   & \multicolumn{1}{c|}{0.935}     & \multicolumn{1}{c|}{0.945}  & 0.935    \\ \cmidrule(l){2-5} 
                      & Unseen & \multicolumn{1}{c|}{0.330}     & \multicolumn{1}{c|}{0.400}  & 0.325    \\ \midrule
\multirow{2}{*}{SVM}  & Seen   & \multicolumn{1}{c|}{0.835}     & \multicolumn{1}{c|}{0.680}  & 0.670    \\ \cmidrule(l){2-5} 
                      & Unseen & \multicolumn{1}{c|}{0.735}     & \multicolumn{1}{c|}{0.665}  & 0.570    \\ \midrule
\multirow{2}{*}{DCNN} & Seen   & \multicolumn{1}{c|}{0.980}     & \multicolumn{1}{c|}{0.985}  & 0.980    \\ \cmidrule(l){2-5} 
                      & Unseen & \multicolumn{1}{c|}{0.535}     & \multicolumn{1}{c|}{0.505}  & 0.385    \\ \midrule
\multirow{2}{*}{LIMU-LSTM} & Seen                          & \multicolumn{1}{c|}{0.960}          & \multicolumn{1}{c|}{0.985}          & 0.975          \\ \cmidrule(l){2-5} 
                      & Unseen & \multicolumn{1}{c|}{0.720}     & \multicolumn{1}{c|}{0.700}  & 0.700    \\ \midrule
GPT4 - DO             & Unseen & \multicolumn{1}{c|}{0.555}     & \multicolumn{1}{c|}{0.570}  & 0.565    \\ \midrule
GPT4 - CoT                 & Unseen                        & \multicolumn{1}{c|}{\textbf{0.790}} & \multicolumn{1}{c|}{\textbf{0.795}} & \textbf{0.790} \\ \bottomrule
\end{tabular}%
}
\caption{Overall test restuls on HHAR dataset.}
\label{tab:res-hhar}
\end{table}

\begin{table}[t]
\resizebox{0.9\columnwidth}{!}{%
\begin{tabular}{@{}c|c|c|c|c|c|c|c@{}}
\toprule
Label                       & Metrics   & RF   & SVM           & DCNN & LIMU-LSTM & GPT4 - DO & GPT4 - CoT    \\ \midrule
\multirow{3}{*}{Downstairs} & Precision & 0.42 & \textbf{1.00} & 0.55 & 0.76      & 0.48      & 0.73          \\ \cmidrule(l){2-8} 
                            & Recall    & 0.72 & 0.33          & 0.05 & 0.57      & 0.55      & \textbf{0.80} \\ \cmidrule(l){2-8} 
                            & F1-score  & 0.53 & 0.50          & 0.09 & 0.65      & 0.51      & \textbf{0.76} \\ \midrule
\multirow{3}{*}{Upstairs}   & Precision & 0.24 & 0.47          & 0.52 & 0.68      & 0.63      & \textbf{0.85} \\ \cmidrule(l){2-8} 
                            & Recall    & 0.08 & \textbf{1.00} & 0.96 & 0.83      & 0.59      & 0.79          \\ \cmidrule(l){2-8} 
                            & F1-score  & 0.12 & 0.64          & 0.68 & 0.75      & 0.62      & \textbf{0.82} \\ \bottomrule
\end{tabular}%
}
\caption{Performance of each action category on HHAR dataset.}
\label{tab:per-hhar}
\end{table}

\begin{figure*}[t]
  \begin{center}
  \includegraphics[width=0.85\textwidth]{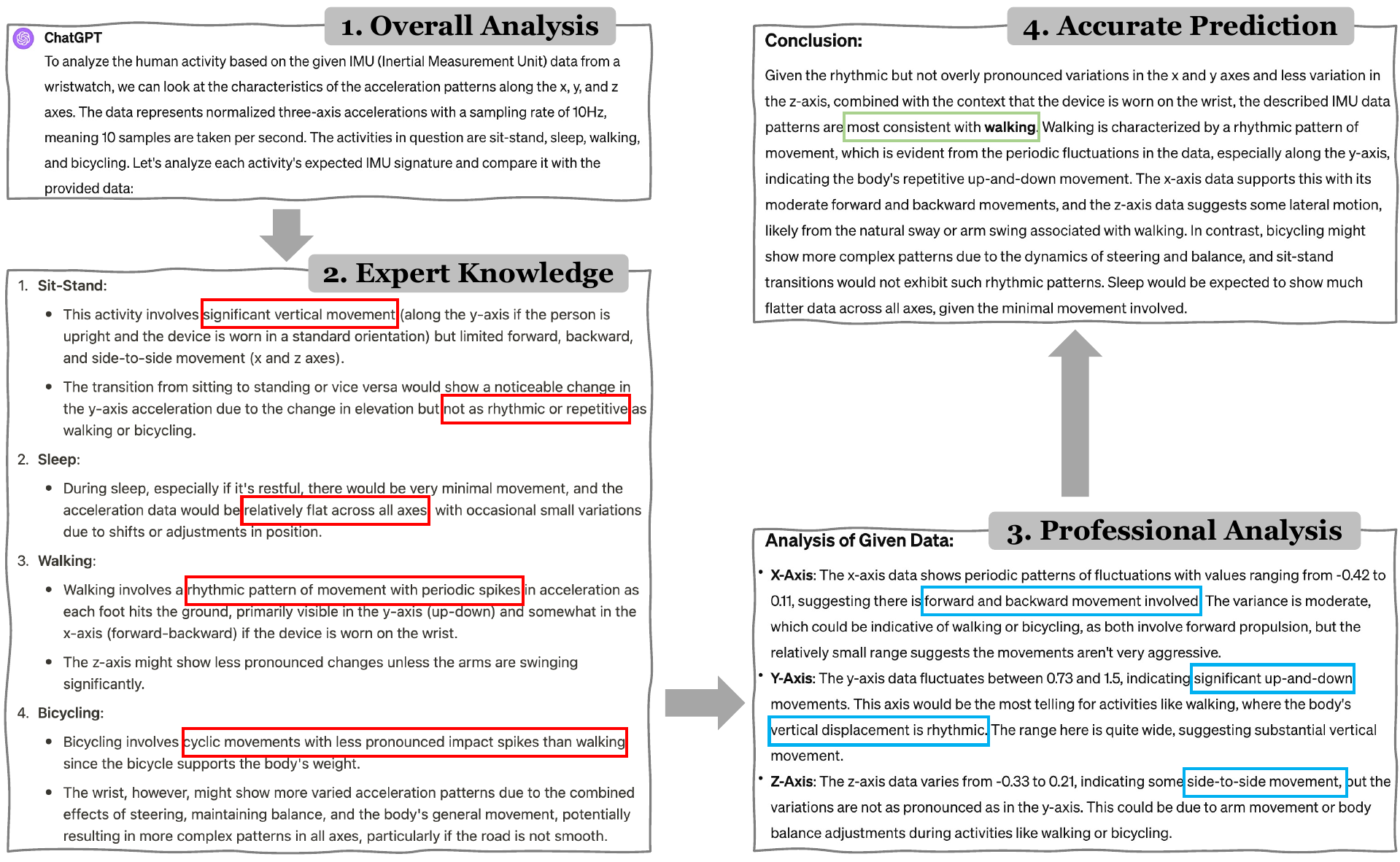}
  \caption{Detailed step-by-step inference generated by GPT4 with a walking example.}\label{fig:gpt4inference}
  \end{center}
\end{figure*}

\head{Detailed Inference Example\footnote{Due to the space limit, to explore more examples, you can visit our Github page:
https://github.com/aiot-lab/HARGPT.}}
To showcase the proficiency of GPT4 in producing expert-level knowledge and precise inference, we present a comprehensive illustration of the concept of walking, as depicted in \fig\ref{fig:gpt4inference}.
The inference process is divided into four parts.
First of all, it explains the information given and makes it clear what problem it needs to solve.
The second step combines hidden "expert knowledge" to accurately characterize the raw IMU data corresponding to each action.
For example, bicycling and walking both have periodic characteristics, sleep is almost non-fluctuating, and sit-stand has sudden changes in value.
In the third installment, GPT4 performs further analysis of the input raw data, which leads to the conclusion that the data is characterized by periodic and up-and-down movement.
After accurate analysis and combining previous expert knowledge, in the fourth part, GPT4 makes an accurate judgment that the movement is most likely walking.

\begin{figure*}[t]
\centering
   \subfloat[GPT3.5]{%
    \label{fig:gpt35}
      \includegraphics[width=0.26\textwidth]{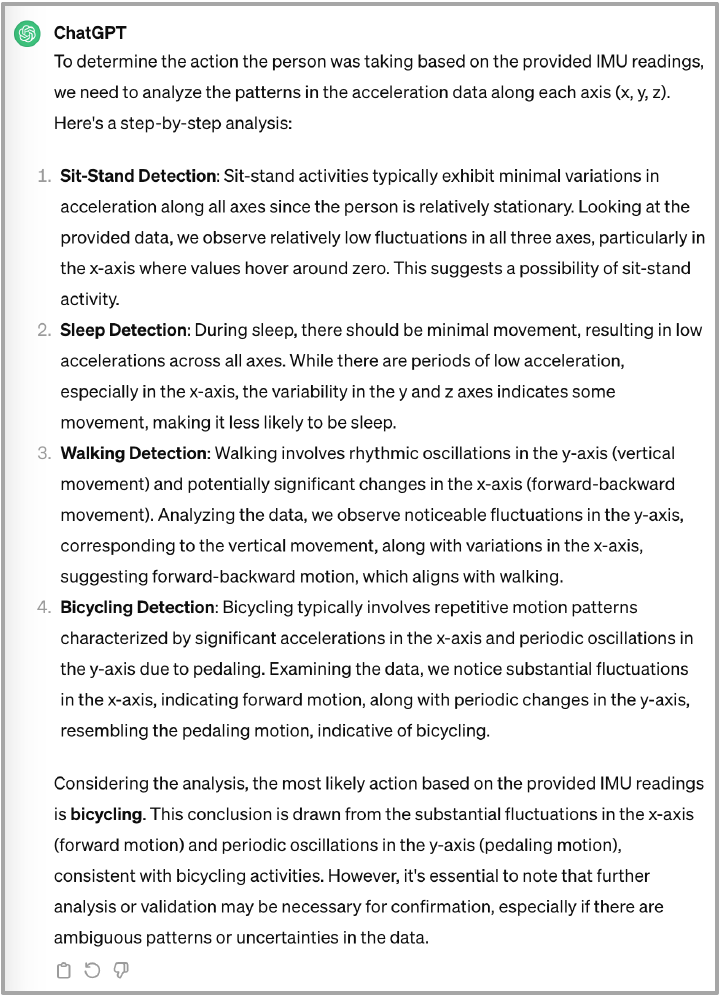}
	  }%
   \subfloat[Gemini]{%
    \label{fig:gemini}
      \includegraphics[width=0.28\textwidth]{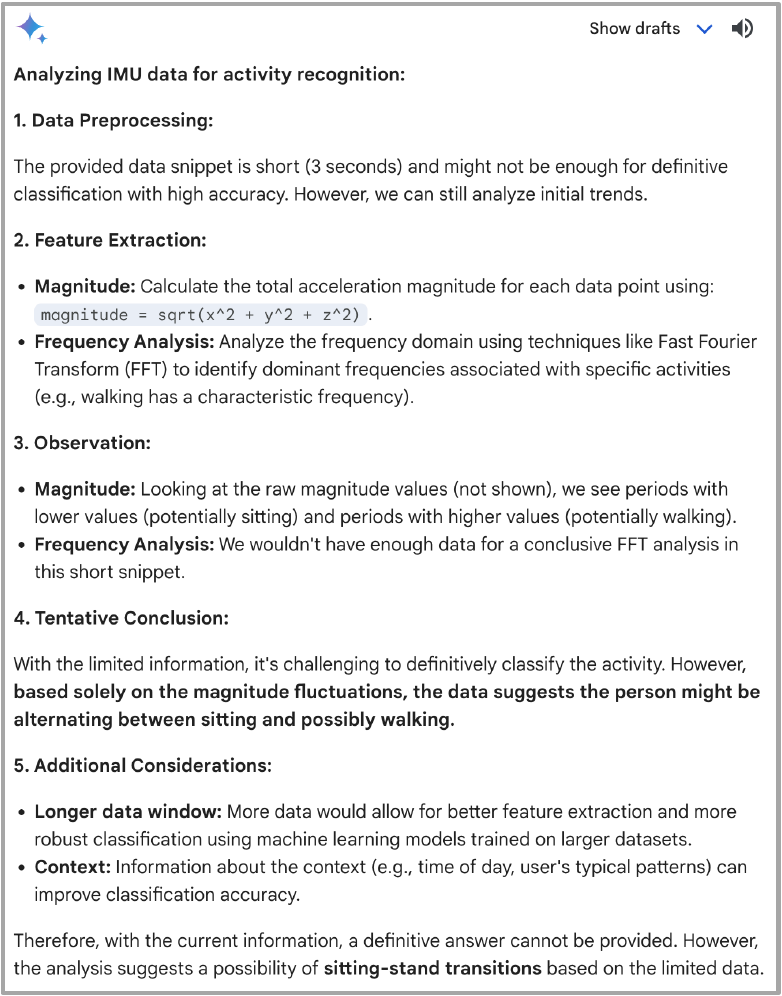}
	  }%
   \subfloat[Llama2-70b]{%
    \label{fig:llama2}
      \includegraphics[width=0.252\textwidth]{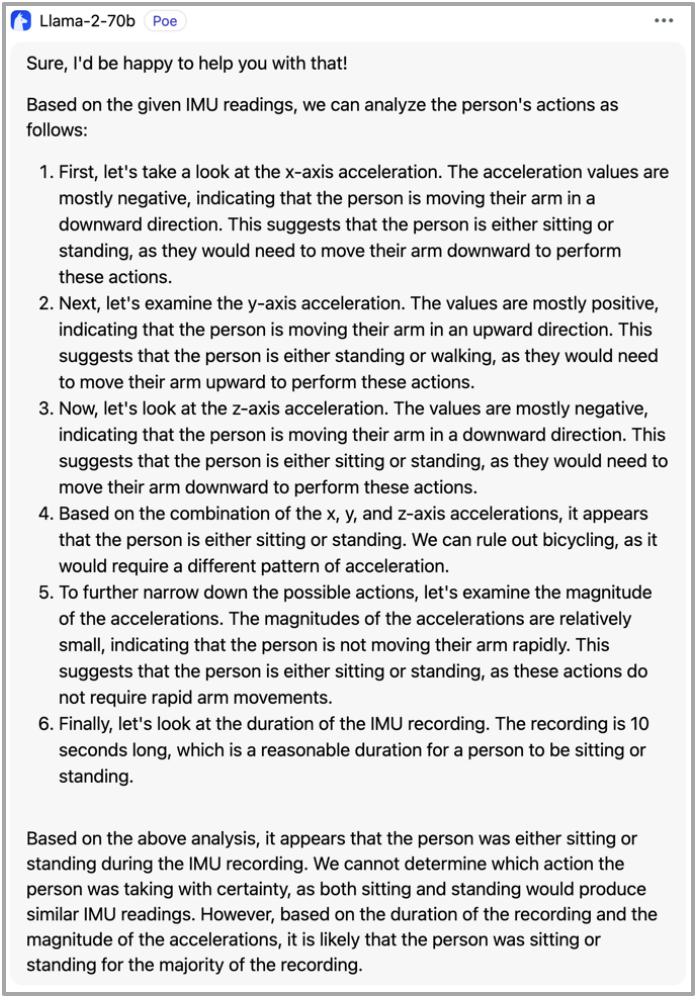}
	  }%
   \caption{A comparison of the inference results generated by other LLMs for the walking scenario.
}
   \label{fig:other-llms}
\end{figure*}

\section{Discussion}
\subsection{Special Properties of LLMs}
\head{Logical Reasoning Ability}
Based on the findings of prior research, it has been established that LLMs possess the capacity for logical reasoning.
Furthermore, our observations indicate that this ability extends beyond textual information processing and encompasses the analysis of data derived from the physical realm.
Existing LLMs such as GPT4 are powerful adapters with the capability to convert unprocessed, raw digitized sensor data into abstract linguistic representations, such as periodic, stationary, abrupt, etc.
This can be likened to an exceptionally potent filter that addresses the challenge of handling out-of-distribution~\cite{liang2017enhancing} data in conventional machine learning or deep learning approaches.
Rather than focusing on learning to match the specific characteristics of raw data, LLMs strive to generate the most plausible assertions by emulating human thought processes.
However, different LLMs vary in their logical abilities. Figure \fig\ref{fig:other-llms} shows the results of providing the same prompt to GPT3.5~\cite{kocon2023chatgpt}, Google Gemini~\cite{team2023gemini}, and Llama2-70b~\cite{touvron2023llama}. We found that LLama2-70b is tentative to provide two possible answers after logical reasoning, instead of a deterministic answer. 

\begin{figure}[t]
  \begin{center}
  \includegraphics[width=0.46\textwidth]{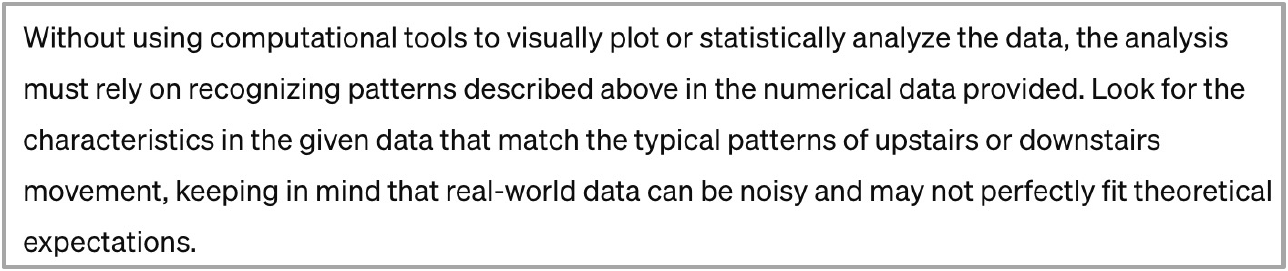}
  \caption{Perfunctory answer generated by GPT4.}\label{fig:perfunctory}
  \end{center}
  \vspace{-0.3cm}
\end{figure}

\head{Perfunctory Answer}
There have been instances where ChatGPT has produced results that are considered perfunctory~\cite{chatgpt2023}.
In our tests, as shown in \fig\ref{fig:perfunctory}.
We have observed that GPT4 occasionally provides constructive opinions without corresponding answers as well. It is crucial to mitigate this issue in order to achieve real-time and stable interaction between LLMs and CPS.

\section{Future Work and Conclusions}
\label{sec:conclusion}
This study demonstrates that LLMs can be used as a foundational model to perform HAR in a zero-shot manner with high accuracy and robustness, which shows that LLMs are capable of processing IoT sensor data without the expertise-guided examples given. The findings highlight the capability of LLMs to analyze the raw sensor data and imply a transformative impact on the Cyber-Physical Systems domain. Nevertheless, further study should be carried out to assess to what extent and in what case the LLMs are effective. Standardizing the evaluation process with a more comprehensive set of evaluation procedures and benchmarks is needed. By deepening our comprehension of the capabilities and constraints of Language Models, we can harness their prowess to advance our comprehension of real-world data, including abstract data like the WiFi Channel State Information.

\bibliographystyle{ACM-Reference-Format}
\bibliography{refs}


\begin{thebibliography}{21}


\ifx \showCODEN    \undefined \def \showCODEN     #1{\unskip}     \fi
\ifx \showDOI      \undefined \def \showDOI       #1{#1}\fi
\ifx \showISBNx    \undefined \def \showISBNx     #1{\unskip}     \fi
\ifx \showISBNxiii \undefined \def \showISBNxiii  #1{\unskip}     \fi
\ifx \showISSN     \undefined \def \showISSN      #1{\unskip}     \fi
\ifx \showLCCN     \undefined \def \showLCCN      #1{\unskip}     \fi
\ifx \shownote     \undefined \def \shownote      #1{#1}          \fi
\ifx \showarticletitle \undefined \def \showarticletitle #1{#1}   \fi
\ifx \showURL      \undefined \def \showURL       {\relax}        \fi
\providecommand\bibfield[2]{#2}
\providecommand\bibinfo[2]{#2}
\providecommand\natexlab[1]{#1}
\providecommand\showeprint[2][]{arXiv:#2}

\bibitem[Achiam et~al\mbox{.}(2023)]%
        {achiam2023gpt}
\bibfield{author}{\bibinfo{person}{Josh Achiam}, \bibinfo{person}{Steven Adler}, \bibinfo{person}{Sandhini Agarwal}, \bibinfo{person}{Lama Ahmad}, \bibinfo{person}{Ilge Akkaya}, \bibinfo{person}{Florencia~Leoni Aleman}, \bibinfo{person}{Diogo Almeida}, \bibinfo{person}{Janko Altenschmidt}, \bibinfo{person}{Sam Altman}, \bibinfo{person}{Shyamal Anadkat}, {et~al\mbox{.}}} \bibinfo{year}{2023}\natexlab{}.
\newblock \showarticletitle{Gpt-4 technical report}.
\newblock \bibinfo{journal}{\emph{arXiv preprint arXiv:2303.08774}} (\bibinfo{year}{2023}).
\newblock


\bibitem[Biau and Scornet(2016)]%
        {biau2016random}
\bibfield{author}{\bibinfo{person}{G{\'e}rard Biau} {and} \bibinfo{person}{Erwan Scornet}.} \bibinfo{year}{2016}\natexlab{}.
\newblock \showarticletitle{A random forest guided tour}.
\newblock \bibinfo{journal}{\emph{Test}}  \bibinfo{volume}{25} (\bibinfo{year}{2016}), \bibinfo{pages}{197--227}.
\newblock


\bibitem[Brooks et~al\mbox{.}(2024)]%
        {videoworldsimulators2024}
\bibfield{author}{\bibinfo{person}{Tim Brooks}, \bibinfo{person}{Bill Peebles}, \bibinfo{person}{Connor Homes}, \bibinfo{person}{Will DePue}, \bibinfo{person}{Yufei Guo}, \bibinfo{person}{Li Jing}, \bibinfo{person}{David Schnurr}, \bibinfo{person}{Joe Taylor}, \bibinfo{person}{Troy Luhman}, \bibinfo{person}{Eric Luhman}, \bibinfo{person}{Clarence Ng}, \bibinfo{person}{Ricky Wang}, {and} \bibinfo{person}{Aditya Ramesh}.} \bibinfo{year}{2024}\natexlab{}.
\newblock \showarticletitle{Video generation models as world simulators}.
\newblock  (\bibinfo{year}{2024}).
\newblock
\urldef\tempurl%
\url{https://openai.com/research/video-generation-models-as-world-simulators}
\showURL{%
\tempurl}


\bibitem[Chan~Chang et~al\mbox{.}(2021)]%
        {chan2021capture}
\bibfield{author}{\bibinfo{person}{S Chan~Chang}, \bibinfo{person}{R Walmsley}, \bibinfo{person}{J Gershuny}, \bibinfo{person}{T Harms}, \bibinfo{person}{E Thomas}, \bibinfo{person}{K Milton}, \bibinfo{person}{P Kelly}, \bibinfo{person}{C Foster}, \bibinfo{person}{A Wong}, \bibinfo{person}{N Gray}, {et~al\mbox{.}}} \bibinfo{year}{2021}\natexlab{}.
\newblock \showarticletitle{Capture-24: Activity tracker dataset for human activity recognition}.
\newblock  (\bibinfo{year}{2021}).
\newblock


\bibitem[ChatGPT(2023)]%
        {chatgpt2023}
\bibfield{author}{\bibinfo{person}{ChatGPT}.} \bibinfo{year}{2023}\natexlab{}.
\newblock \bibinfo{booktitle}{}.
\newblock
\urldef\tempurl%
\url{https://twitter.com/ChatGPTapp/status/1732979491071549792}
\showURL{%
\tempurl}


\bibitem[Hearst et~al\mbox{.}(1998)]%
        {hearst1998support}
\bibfield{author}{\bibinfo{person}{Marti~A. Hearst}, \bibinfo{person}{Susan~T Dumais}, \bibinfo{person}{Edgar Osuna}, \bibinfo{person}{John Platt}, {and} \bibinfo{person}{Bernhard Scholkopf}.} \bibinfo{year}{1998}\natexlab{}.
\newblock \showarticletitle{Support vector machines}.
\newblock \bibinfo{journal}{\emph{IEEE Intelligent Systems and their applications}} \bibinfo{volume}{13}, \bibinfo{number}{4} (\bibinfo{year}{1998}), \bibinfo{pages}{18--28}.
\newblock


\bibitem[Kim et~al\mbox{.}(2024)]%
        {kim2024health}
\bibfield{author}{\bibinfo{person}{Yubin Kim}, \bibinfo{person}{Xuhai Xu}, \bibinfo{person}{Daniel McDuff}, \bibinfo{person}{Cynthia Breazeal}, {and} \bibinfo{person}{Hae~Won Park}.} \bibinfo{year}{2024}\natexlab{}.
\newblock \showarticletitle{Health-llm: Large language models for health prediction via wearable sensor data}.
\newblock \bibinfo{journal}{\emph{arXiv preprint arXiv:2401.06866}} (\bibinfo{year}{2024}).
\newblock


\bibitem[Koco{\'n} et~al\mbox{.}(2023)]%
        {kocon2023chatgpt}
\bibfield{author}{\bibinfo{person}{Jan Koco{\'n}}, \bibinfo{person}{Igor Cichecki}, \bibinfo{person}{Oliwier Kaszyca}, \bibinfo{person}{Mateusz Kochanek}, \bibinfo{person}{Dominika Szyd{\l}o}, \bibinfo{person}{Joanna Baran}, \bibinfo{person}{Julita Bielaniewicz}, \bibinfo{person}{Marcin Gruza}, \bibinfo{person}{Arkadiusz Janz}, \bibinfo{person}{Kamil Kanclerz}, {et~al\mbox{.}}} \bibinfo{year}{2023}\natexlab{}.
\newblock \showarticletitle{ChatGPT: Jack of all trades, master of none}.
\newblock \bibinfo{journal}{\emph{Information Fusion}} (\bibinfo{year}{2023}), \bibinfo{pages}{101861}.
\newblock


\bibitem[Kojima et~al\mbox{.}(2022)]%
        {kojima2022large}
\bibfield{author}{\bibinfo{person}{Takeshi Kojima}, \bibinfo{person}{Shixiang~Shane Gu}, \bibinfo{person}{Machel Reid}, \bibinfo{person}{Yutaka Matsuo}, {and} \bibinfo{person}{Yusuke Iwasawa}.} \bibinfo{year}{2022}\natexlab{}.
\newblock \showarticletitle{Large language models are zero-shot reasoners}.
\newblock \bibinfo{journal}{\emph{Advances in neural information processing systems}}  \bibinfo{volume}{35} (\bibinfo{year}{2022}), \bibinfo{pages}{22199--22213}.
\newblock


\bibitem[LeCun(2022)]%
        {lecun2022path}
\bibfield{author}{\bibinfo{person}{Yann LeCun}.} \bibinfo{year}{2022}\natexlab{}.
\newblock \showarticletitle{A path towards autonomous machine intelligence version 0.9. 2, 2022-06-27}.
\newblock \bibinfo{journal}{\emph{Open Review}}  \bibinfo{volume}{62} (\bibinfo{year}{2022}).
\newblock


\bibitem[Li et~al\mbox{.}(2024)]%
        {li2024personal}
\bibfield{author}{\bibinfo{person}{Yuanchun Li}, \bibinfo{person}{Hao Wen}, \bibinfo{person}{Weijun Wang}, \bibinfo{person}{Xiangyu Li}, \bibinfo{person}{Yizhen Yuan}, \bibinfo{person}{Guohong Liu}, \bibinfo{person}{Jiacheng Liu}, \bibinfo{person}{Wenxing Xu}, \bibinfo{person}{Xiang Wang}, \bibinfo{person}{Yi Sun}, {et~al\mbox{.}}} \bibinfo{year}{2024}\natexlab{}.
\newblock \showarticletitle{Personal llm agents: Insights and survey about the capability, efficiency and security}.
\newblock \bibinfo{journal}{\emph{arXiv preprint arXiv:2401.05459}} (\bibinfo{year}{2024}).
\newblock


\bibitem[Liang et~al\mbox{.}(2017)]%
        {liang2017enhancing}
\bibfield{author}{\bibinfo{person}{Shiyu Liang}, \bibinfo{person}{Yixuan Li}, {and} \bibinfo{person}{Rayadurgam Srikant}.} \bibinfo{year}{2017}\natexlab{}.
\newblock \showarticletitle{Enhancing the reliability of out-of-distribution image detection in neural networks}.
\newblock \bibinfo{journal}{\emph{arXiv preprint arXiv:1706.02690}} (\bibinfo{year}{2017}).
\newblock


\bibitem[Nori et~al\mbox{.}(2023)]%
        {nori2023can}
\bibfield{author}{\bibinfo{person}{Harsha Nori}, \bibinfo{person}{Yin~Tat Lee}, \bibinfo{person}{Sheng Zhang}, \bibinfo{person}{Dean Carignan}, \bibinfo{person}{Richard Edgar}, \bibinfo{person}{Nicolo Fusi}, \bibinfo{person}{Nicholas King}, \bibinfo{person}{Jonathan Larson}, \bibinfo{person}{Yuanzhi Li}, \bibinfo{person}{Weishung Liu}, {et~al\mbox{.}}} \bibinfo{year}{2023}\natexlab{}.
\newblock \showarticletitle{Can generalist foundation models outcompete special-purpose tuning? case study in medicine}.
\newblock \bibinfo{journal}{\emph{arXiv preprint arXiv:2311.16452}} (\bibinfo{year}{2023}).
\newblock


\bibitem[Pedregosa et~al\mbox{.}(2011)]%
        {scikit-learn}
\bibfield{author}{\bibinfo{person}{F. Pedregosa}, \bibinfo{person}{G. Varoquaux}, \bibinfo{person}{A. Gramfort}, \bibinfo{person}{V. Michel}, \bibinfo{person}{B. Thirion}, \bibinfo{person}{O. Grisel}, \bibinfo{person}{M. Blondel}, \bibinfo{person}{P. Prettenhofer}, \bibinfo{person}{R. Weiss}, \bibinfo{person}{V. Dubourg}, \bibinfo{person}{J. Vanderplas}, \bibinfo{person}{A. Passos}, \bibinfo{person}{D. Cournapeau}, \bibinfo{person}{M. Brucher}, \bibinfo{person}{M. Perrot}, {and} \bibinfo{person}{E. Duchesnay}.} \bibinfo{year}{2011}\natexlab{}.
\newblock \showarticletitle{Scikit-learn: Machine Learning in {P}ython}.
\newblock \bibinfo{journal}{\emph{Journal of Machine Learning Research}}  \bibinfo{volume}{12} (\bibinfo{year}{2011}), \bibinfo{pages}{2825--2830}.
\newblock


\bibitem[Stisen et~al\mbox{.}(2015)]%
        {stisen2015smart}
\bibfield{author}{\bibinfo{person}{Allan Stisen}, \bibinfo{person}{Henrik Blunck}, \bibinfo{person}{Sourav Bhattacharya}, \bibinfo{person}{Thor~Siiger Prentow}, \bibinfo{person}{Mikkel~Baun Kj{\ae}rgaard}, \bibinfo{person}{Anind Dey}, \bibinfo{person}{Tobias Sonne}, {and} \bibinfo{person}{Mads~M{\o}ller Jensen}.} \bibinfo{year}{2015}\natexlab{}.
\newblock \showarticletitle{Smart devices are different: Assessing and mitigatingmobile sensing heterogeneities for activity recognition}. In \bibinfo{booktitle}{\emph{Proceedings of the 13th ACM conference on embedded networked sensor systems}}. \bibinfo{pages}{127--140}.
\newblock


\bibitem[Team et~al\mbox{.}(2023)]%
        {team2023gemini}
\bibfield{author}{\bibinfo{person}{Gemini Team}, \bibinfo{person}{Rohan Anil}, \bibinfo{person}{Sebastian Borgeaud}, \bibinfo{person}{Yonghui Wu}, \bibinfo{person}{Jean-Baptiste Alayrac}, \bibinfo{person}{Jiahui Yu}, \bibinfo{person}{Radu Soricut}, \bibinfo{person}{Johan Schalkwyk}, \bibinfo{person}{Andrew~M Dai}, \bibinfo{person}{Anja Hauth}, {et~al\mbox{.}}} \bibinfo{year}{2023}\natexlab{}.
\newblock \showarticletitle{Gemini: a family of highly capable multimodal models}.
\newblock \bibinfo{journal}{\emph{arXiv preprint arXiv:2312.11805}} (\bibinfo{year}{2023}).
\newblock


\bibitem[Touvron et~al\mbox{.}(2023)]%
        {touvron2023llama}
\bibfield{author}{\bibinfo{person}{Hugo Touvron}, \bibinfo{person}{Louis Martin}, \bibinfo{person}{Kevin Stone}, \bibinfo{person}{Peter Albert}, \bibinfo{person}{Amjad Almahairi}, \bibinfo{person}{Yasmine Babaei}, \bibinfo{person}{Nikolay Bashlykov}, \bibinfo{person}{Soumya Batra}, \bibinfo{person}{Prajjwal Bhargava}, \bibinfo{person}{Shruti Bhosale}, {et~al\mbox{.}}} \bibinfo{year}{2023}\natexlab{}.
\newblock \showarticletitle{Llama 2: Open foundation and fine-tuned chat models}.
\newblock \bibinfo{journal}{\emph{arXiv preprint arXiv:2307.09288}} (\bibinfo{year}{2023}).
\newblock


\bibitem[Wei et~al\mbox{.}(2022)]%
        {wei2022chain}
\bibfield{author}{\bibinfo{person}{Jason Wei}, \bibinfo{person}{Xuezhi Wang}, \bibinfo{person}{Dale Schuurmans}, \bibinfo{person}{Maarten Bosma}, \bibinfo{person}{Fei Xia}, \bibinfo{person}{Ed Chi}, \bibinfo{person}{Quoc~V Le}, \bibinfo{person}{Denny Zhou}, {et~al\mbox{.}}} \bibinfo{year}{2022}\natexlab{}.
\newblock \showarticletitle{Chain-of-thought prompting elicits reasoning in large language models}.
\newblock \bibinfo{journal}{\emph{Advances in Neural Information Processing Systems}}  \bibinfo{volume}{35} (\bibinfo{year}{2022}), \bibinfo{pages}{24824--24837}.
\newblock


\bibitem[Xu et~al\mbox{.}(2024)]%
        {xu2024penetrative}
\bibfield{author}{\bibinfo{person}{Huatao Xu}, \bibinfo{person}{Liying Han}, \bibinfo{person}{Qirui Yang}, \bibinfo{person}{Mo Li}, {and} \bibinfo{person}{Mani Srivastava}.} \bibinfo{year}{2024}\natexlab{}.
\newblock \showarticletitle{Penetrative AI: Making LLMs Comprehend the Physical World}. In \bibinfo{booktitle}{\emph{Proceedings of the 25th International Workshop on Mobile Computing Systems and Applications}}. \bibinfo{pages}{1--7}.
\newblock


\bibitem[Xu et~al\mbox{.}(2022)]%
        {xu2022limu}
\bibfield{author}{\bibinfo{person}{Huatao Xu}, \bibinfo{person}{Pengfei Zhou}, \bibinfo{person}{Rui Tan}, \bibinfo{person}{Mo Li}, {and} \bibinfo{person}{Guobin Shen}.} \bibinfo{year}{2022}\natexlab{}.
\newblock \showarticletitle{LIMU-BERT: Unleashing the Potential of Unlabeled Data for IMU Sensing Applications}.
\newblock \bibinfo{journal}{\emph{GetMobile: Mobile Computing and Communications}} \bibinfo{volume}{26}, \bibinfo{number}{3} (\bibinfo{year}{2022}), \bibinfo{pages}{39--42}.
\newblock


\bibitem[Yang et~al\mbox{.}(2015)]%
        {yang2015deep}
\bibfield{author}{\bibinfo{person}{Jianbo Yang}, \bibinfo{person}{Minh~Nhut Nguyen}, \bibinfo{person}{Phyo~Phyo San}, \bibinfo{person}{Xiaoli Li}, {and} \bibinfo{person}{Shonali Krishnaswamy}.} \bibinfo{year}{2015}\natexlab{}.
\newblock \showarticletitle{Deep convolutional neural networks on multichannel time series for human activity recognition.}. In \bibinfo{booktitle}{\emph{Ijcai}}, Vol.~\bibinfo{volume}{15}. Buenos Aires, Argentina, \bibinfo{pages}{3995--4001}.
\newblock


\end{thebibliography}

\end{document}